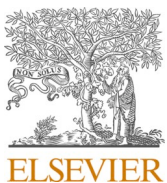
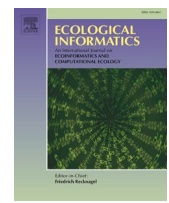

# Metadata augmented deep neural networks for wild animal classification

Aslak Tøn [a], Ammar Ahmed [a], Ali Shariq Imran [a,*], Mohib Ullah [a], R. Muhammad Atif Azad [b]

[a] *Intelligent Systems and Analytics (ISA) Research Group, Department of Computer Science, Norwegian University of Science & Technology (NTNU), Gjøvik 2815, Norway*
[b] *College of Computing, Faculty of Computing, Engineering and Built Environment, Birmingham City University., Birmingham, United Kingdom*

ABSTRACT

Camera trap imagery has become an invaluable asset in contemporary wildlife surveillance, enabling researchers to observe and investigate the behaviors of wild animals. While existing methods rely solely on image data for classification, this may not suffice in cases of suboptimal animal angles, lighting, or image quality. This study introduces a novel approach that enhances wild animal classification by combining specific metadata (temperature, location, time, etc) with image data. Using a dataset focused on the Norwegian climate, our models show an accuracy increase from 98.4% to 98.9% compared to existing methods. Notably, our approach also achieves high accuracy with metadata-only classification, highlighting its potential to reduce reliance on image quality. This work paves the way for integrated systems that advance wildlife classification technology.

## 1. Introduction

Over the past few centuries, human expansion and exploitation of natural resources have significantly stressed global wildlife. Factors such as human-induced climate change (Pörtner et al., 2022), deforestation (Lata et al., 2018), and the proliferation of roads (Richard et al., 2019) have led to a decline in biodiversity, marking a modern mass extinction event (Pievani, 2014). This pervasive impact has defined a new geological era, the Anthropocene, characterized by the profound influence of human activities on the planet. Monitoring wild habitats is imperative to understand and mitigate these impacts, providing invaluable data for informed ecosystem management. Camera traps have become a vital tool in wildlife surveillance, allowing for extensive data collection with minimal disruption to natural habitats. This data is crucial for monitoring animal populations, biodiversity, and individual behaviors. However, the primary challenge lies in extracting relevant information from the images, which requires tagging, labeling, and sorting. Citizen science has been instrumental in addressing this labor-intensive task, but the growing volume of data necessitates more efficient solutions (Swanson et al., 2015).

Deep learning techniques have emerged as a powerful solution for wildlife image classification, leveraging large datasets to achieve high detection rates. While early contributions used pattern matching (Bolger et al., 2012) or feature extraction followed by classification via support vector machines (Yu et al., 2013), Chen et al. (2014) introduced the use of convolutional neural networks (CNNs) and an early form of object detection to wildlife camera trap literature. Gomez Villa et al. (2017) introduced transfer learning to improve the performance of CNN classification. The use of deep learning methods for the automatic classification of wildlife camera trap images has become widespread in recent years, with the work of Norouzzadeh et al. (2018) being another significant contribution in the field, in which they used deep convolutional neural networks to detect, identify, and count wild animal species. Subsequently, several research groups incorporated an object detection component, e.g., (Norouzzadeh et al., 2021; Shepley et al., 2021; Tabak et al., 2020), or discussed out-of-sample performance, e.g., (Auer et al., 2021; Curry et al., 2021; Gimenez et al., 2021; Miao et al., 2021; Schneider et al., 2019; Shepley et al., 2021; Tabak et al., 2020; Whytock et al., 2021). However, this field is not without its challenges. Wild animals seldom strike photogenic poses, and achieving ideal lighting conditions can be elusive, particularly for nocturnal species. Nevertheless, when the models are specifically tailored for Wild Animal Classification (WAC), deep learning emerges as a powerful tool, significantly easing the workload of scientists. This enables them to shift their focus from image labeling to more in-depth data analysis, made feasible by automatic WAC. While the domain of automatic WAC has been extensively explored in datasets like Snapshot Serengeti, applying models trained on one dataset to new data from different climatic regions, such as Norway, results in notable declines in accuracy due to marked differences in biodiversity and landscapes. Therefore, deep learning models trained on the Snapshot dataset may not perform optimally in the Norwegian climate.

Additionally, camera trap images vary in lighting conditions, with RGB daytime and IR nighttime captures complicating model training. Simple grayscale conversion fails to capture the nuances of these lighting variations, necessitating models that can handle both grayscale and color images. Moreover, suboptimal image angles and varying image quality can further hinder classification accuracy. To address these limitations, our study introduces a novel approach that integrates specific metadata (temperature, location, time) with image data, enhancing classification performance. This method is particularly effective in scenarios with suboptimal lighting or image quality, reducing reliance on image data alone. Using a dataset from the Norwegian Institute for Nature Research (NINA) (Nina, n.d.), we developed region-specific models tailored to Norway's unique environmental conditions. Our approach not only improves classification accuracy but also achieves high accuracy with metadata-only classification, highlighting its potential to overcome image quality issues.

This paper presents the following primary contributions:

- We have curated an extensive dataset of 170,000 image and metadata samples from the NINA Viltkamera dataset. We also fill in the missing metadata when possible or provide contextual information to the deep learning network when data is not valid.
- Our study showcases the potential of fusing metadata along with images to significantly improve the performance of WAC models, representing a valuable advancement within the field.
- We investigate and assess diverse data fusion techniques, focusing on integrating image data and metadata. Our findings highlight the advantages of certain methods while shedding light on their limitations.
- We propose a method to logically aggregate various species into supergroups to enhance the model performance.
- We show what types of metadata exert the most influence on WAC.
- We outline effective strategies for incorporating metadata into WAC tasks without increasing the expected annotation work done by experts.

The remainder of the article is structured as follows: Section 2 reviews current literature on wildlife classification. Section 3 details our methodology, covering dataset acquisition and our metadata-augmented models. Section 4 presents the results and discussion. Finally, Section 5 concludes the paper.

## 2. Related work

This section concisely overviews pertinent research and discusses the contemporary techniques employed in classifying wild animals, particularly from the last three years. The first subsection delves into contemporary detection techniques utilized in the realm of wild animal recognition, whereas the second subsection focuses on classification techniques. The last subsection discusses the data fusion in various fields and which inclusion techniques have worked better.

### 2.1. Detection techniques

Saxena et al. (2020) introduced animal detection using SSD and Faster R-CNN on a dataset of 25 animal classes with 31,774 images. They achieved an mAP of 80.5% at 100 fps with SSD and 82.11% at 10 fps with Faster R-CNN. Tan et al. (2022) created the NTLNP wildlife dataset and evaluated several object detection models on it, including YOLOv5m, Cascade R-CNN, and FCOS. YOLOv5m achieved the highest accuracy at 98.90% with the standard mAP (mean average precision) threshold of 0.5. Simões et al. (2023) employed a three-step process: video-to-image conversion, annotation using MegaDetector, and enhanced detection and classification with Inception-ResNetv2-based Faster R-CNN. They achieved 73.92% mAP for classification and 96.88% mAP for detection at an IoU of 0.5. Norouzzadeh et al. (2021) employed an active learning system to reduce the manual effort necessary for training a computer vision model. Additionally, they incorporated object detection models and transfer learning to mitigate the risk of overfitting to specific camera locations; their target datasets were Snapshot Serengeti and NACTI (Lilawp., 2023). This method resulted in an accuracy of 91.71%, a precision of 84.47%, and a recall of 84.24%. Schindler and Steinhage (2021) presented a two-stage fusion network for animal classification, action recognition, and segmentation. They used Mask R-CNN and incorporated temporal data from 528 nighttime video clips involving deer, boars, foxes, and hares, achieving 63.8% average precision (AP) for animal detection and identification and 94.10% accuracy for action detection. Buehler et al. (2019) proposed a method for automatic wild giraffe cropping based on Histogram of Oriented Gradients (HOG) features. They subsequently trained a Support Vector Machine (SVM) classifier using positive and negative HOG descriptors, along with hard-negatives mined through an Active Learning approach. The trained SVM was then utilized to detect occurrences of the object in new images. This process involved sliding a rectangular window over the image and evaluating the trained SVM at each window position to identify all objects. They reported a mean failure rate of 0.109 for 3518 raw photos and a mean failure rate of 0.006 for high-quality photos. Gomez Villa et al. (2017) employed a transfer learning approach in their work. They utilized the Snapshot Serengeti dataset, selecting the most common 26 out of 48 species, to perform automatic classification of animal species in camera-trap images. Their fine-tuned ResNet-101 model achieved the best performance among all eight architectures employed. Qi Song et al. (2024) constructed a dataset comprising 15 bird species using Camera Traps. They employed deep learning techniques to identify birds amidst complex backgrounds, with Cascade RCNN models outperforming other object detection models. Additionally, their study found that the choice of backbones significantly influenced bird recognition accuracy. Recently, Bothmann et al. (2023) provided a comprehensive tuning procedure for optimizing the hyperparameters of a multi-step pipeline, comprising object detection and image classification. Additionally, they introduced an active learning component to facilitate efficient training of a high-performing model on new data, including potential scenarios involving new monitoring locations or previously unseen animal species. Notably, they achieved the highest accuracy of 98.7% with their approach on a dataset featuring five species.

### 2.2. Classification techniques

Dhillon and Verma (2022) introduced a feature extraction and fusion method using DenseNet201 and ResNet101. They reduced feature dimensionality with Neighborhood Component Analysis (NCA), followed by concatenation and SVM for classification. Their approach achieved 98.07% accuracy on a camera trap wild animal dataset. Battu (2022) employed two networks, one with clean samples and one without, using Snapshot Serengeti and Panama-Netherlands datasets. They enhanced training by grouping data via k-means clustering and achieved 73.09% accuracy at a 30% noise level. They observed a decline in the accuracy beyond that noise level. Islam et al. (2023) used DL models for categorizing snakes, lizards, and toads from camera trap images. Their self-trained CNN achieved 72% accuracy, while VGG16 and ResNet50 reached 87% and 86% accuracy, respectively, in multi-class classification. Sreedevi and Edison (2022) designed an algorithm for wild animal detection using a depth-wise separable convolution layer. Their model, which used zero padding to preserve edge characteristics, was tested on the IWildCam dataset, achieving an IoU of 87.8% for detection accuracy and 99.6% accuracy in wild animal classification.

Observing recent literature reveals a prevailing trend where the majority of studies exclusively utilize image data for wild animal classification. However, our study takes a novel approach by incorporating metadata in conjunction with image data. We aim to demonstrate the enhanced potential achieved through the fusion of metadata and image data, thereby advancing the current state of the art.

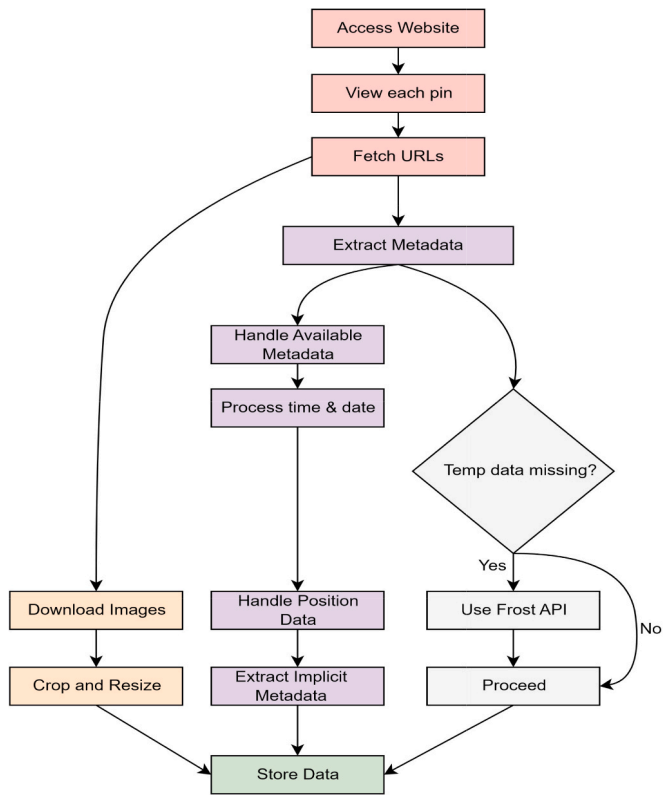

Fig. 1. Process for acquiring the NINA dataset

### 2.3. Data fusion for deep learning

The exploration of data fusion and deep learning has been undertaken by numerous researchers across various fields. The overarching objective of data fusion is to provide additional context to enhance the decision-making capabilities of models. Arevalo et al. (2017) propose a specialized multimodal unit for handling data fusion, termed the Gated Multimodal Unit (GMU). In their approach, they address scenarios involving two types of multimodal data, wherein the contribution of each data type is determined by either $\sigma$ or $1 - \sigma$, with $\sigma$ being a hyperparameter learned during model training. Employing their GMU, they combine textual descriptors with images to predict the genre of movies.

Data fusion has also been applied to skin lesion classification. Pacheco and Krohling (2021) and Li et al. (2020) both introduce approaches for integrating patient data with image data to enhance prediction accuracy. In general, the findings from both studies indicate that early fusion methods outperform late fusion strategies. These papers suggest that incorporating metadata enhances the feature extraction process, thereby improving classification performance.

Finally, Bi et al. (2022) propose the use of user-generated hintmaps along with data fusion to enhance skin lesion segmentation. Following an initial feature extraction stage, positive and negative hintmaps are integrated with the image using multiple Hyper Integration Modules to refine segmentation results. Although the direct application of their model may not be applicable to the image and metadata scenario studied in this paper, their contributions offer valuable insights for designing architectures aimed at predicting animals in camera traps.

## 3. Materials and methods

This section outlines our methodology, beginning with the steps involved in acquiring the image dataset and metadata. Subsequently, we delve into the baseline models, ablation study, and our metadata-augmented models. The section concludes by providing details on training and evaluation processes.

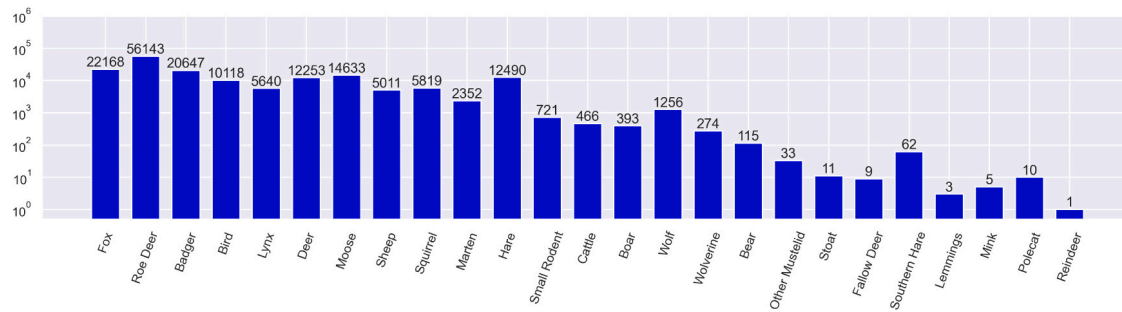

(a) Mild grouping of animals

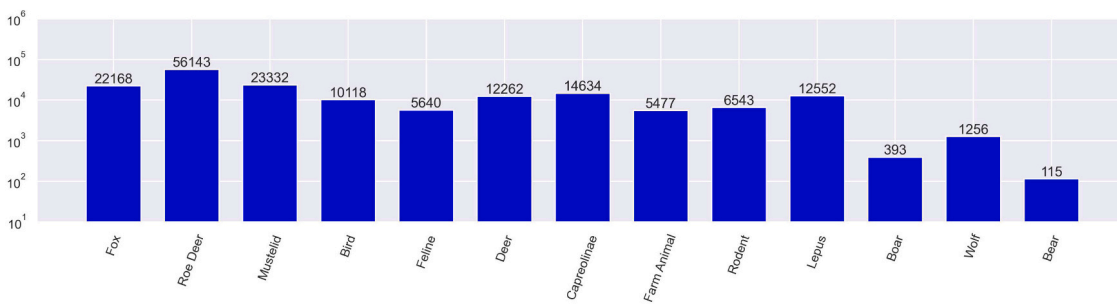

(b) Aggressive grouping of animals

Fig. 2. Data distribution

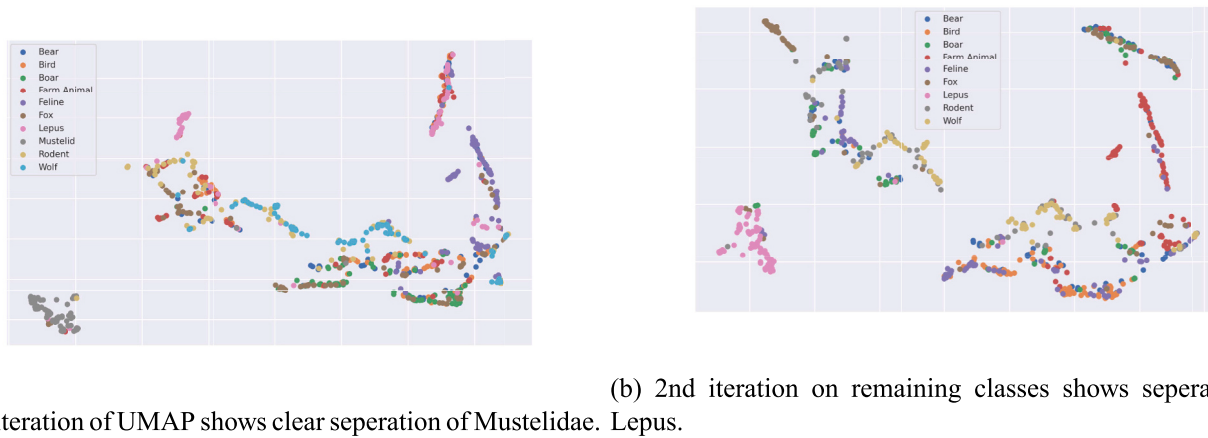

(a) 1st iteration of UMAP shows clear seperation of Mustelidae.

(b) 2nd iteration on remaining classes shows seperation of Lepus.

**Fig. 3.** UMAP projections demonstrating a separation between classes

### 3.1. NINA viltkamera dataset acquisition

We acquired a comprehensive dataset of 170,000 sample images along with metadata from the NINA website[6]. This dataset encompasses images of various animal species within the Norwegian climate, primarily concentrated in the central region of Norway. The dataset acquisition process is shown in Fig. 1.

However, obtaining the NINA Viltkamera dataset was a complex challenge. Initially, we aimed to obtain a downloadable data blob from NINA. However, creating this proved challenging for them, leading to the agreement that a web scraping approach would be more feasible. The NINA website features an interactive map with pins representing individual camera traps, and clicking on these pins allows access to the images captured by each trap. To extract the image URLs, we had to identify the location of these URLs within the website structure. We found that they were stored within an object referred to as "vm" on the website. Further analysis revealed that this object contained a data member named "vm.media", which in turn housed a JSON object containing valuable metadata, including filenames and foreign keys referencing species IDs (NOR: "FK_ArtID"). To establish the link between "FK_ArtID" and the species name, we utilized the "vm.arter()" function. This function returned a list of JSON objects, each containing an "ArtID" and its corresponding species name. Similarly, to pinpoint the approximate camera location (Location ID, latitude, and longitude), we employed the "vm.lokaliteter()" function. We iterated through each pin on the map to access all samples and their respective image URLs. Once we acquired the information, including image filenames, we downloaded the images directly from the provided URLs. To minimize the load on NINA's servers, we conducted image processing tasks, which involved cropping out information bands at the top and bottom of each image. Subsequently, we resized the images to dimensions of 512 × 512.

The NINA Viltkamera dataset encompassed a total of 100 distinct classes for categorizing images. Nevertheless, it is noteworthy that not all of these 100 classes contained any samples. Only 65 classes within the downloaded dataset had one or more samples associated with them. Developing an effective classifier for all 65 classes would present a considerable challenge, particularly given the limited number of samples available for some of these classes. Consequently, we opted to aggregate these classes into broader super-classes. A visual representation of a relatively mild aggregation can be observed in Fig. 2a. The most significant combination here is combining all birds into one super-class "Bird". This 25-class dataset is still quite imbalanced, and while the larger classes like "Roe Deer" will fare well, we worry some of the smaller classes will be ignored. Therefore, we created a more aggressive grouping of the classes, which can be seen in Fig. 2b. To achieve this, we initially divide the dataset into a binary: "Deer" vs. "Not Deer". Subsequently, the "Deer" class is further subdivided into three subclasses: "Roe Deer", "Deer", and "Capreolinae". Hence, we utilized a final dataset of 13 classes in our main experiments involving metadata-augmented neural networks and baseline methods using image and metadata together for classification.

Next, we want to investigate whether its possible to separate out the classes that are most easily distinguishable by the metadata. Our dataset comprises 538 data points, indicating a potential mapping in a 538-dimensional space to discern underlying groupings. Given the challenge of visualizing information beyond three dimensions (or four with temporal data), we turn to dimensionality reduction techniques. Our study employs a novel approach to dimensionality reduction introduced by McInnes et al. (2020). Uniform Manifold Approximation and Projection or UMAP leverages topology, higher-dimensional manifolds, and graph theory to project high-dimensional data into lower dimensions while minimizing cross-entropy between the original and re-projected data. McInnes et al. (2020) demonstrates the qualitative and quantitative superiority of UMAP over several other dimensionality reduction algorithms like t-SNE, LargeVis, Laplacian Eigenmaps, and PCA. The underlying math of the method relies on a good understanding of topology. However, the actual algorithm can be summarized as given by Afridi et al. (2022); Tøn et al. (2023). They break down the process into two major steps and a couple minor steps in each major step as so:

1. Learn manifold structure
   1.1. Finding nearest neighbors
   1.2. Constructing neighbors graph
      1.2.1. Varying distance
      1.2.2. Local connectivity
      1.2.3. Fuzzy area
      1.2.4. Merging of edges
2. Finding low-dimensional representation
   2.1. Minimum distance
   2.2. Minimizing the cost function

By employing UMAP, we can explore potential patterns within animal clusters. If we identify local clusters in the reduced-dimensional space, it suggests that similar patterns likely exist in the original 538-dimensional space, which is otherwise challenging to analyze. Fig. 3a shows a clear separation of the Mustelidae class. This could be because Mustelidae encompasses a distinct group of carnivorous mammals, including species like weasels, otters, and badgers. These species may

---

[6] https://viltkamera.nina.no/

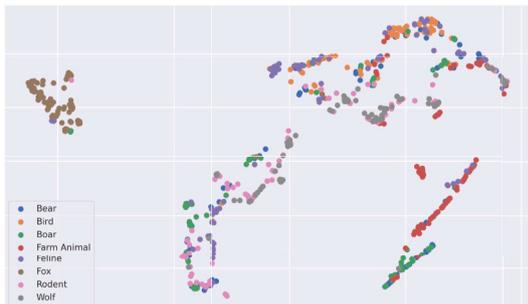

(a) 3rd iteration on remaining data shows seperation of Fox.

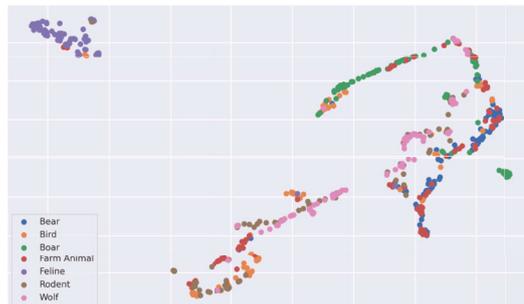

(b) 4th iteration shows seperation of Feline.

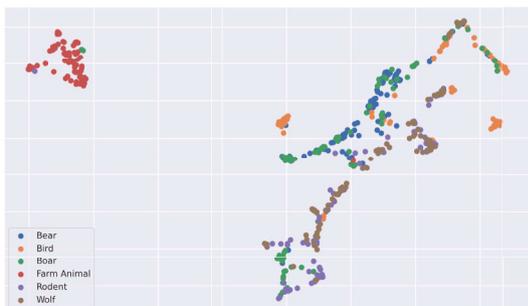

(c) 5th iteration shows seperation of Farm Animal.

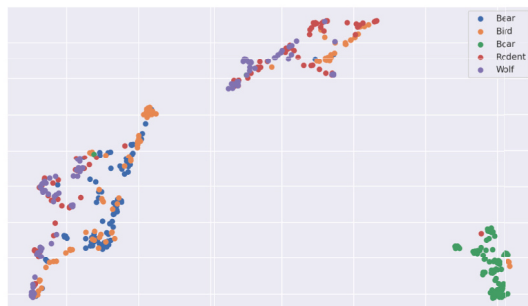

(d) 6th iteration shows seperation of Boar.

**Fig. 4.** UMAP projections demonstrating a separation between classes

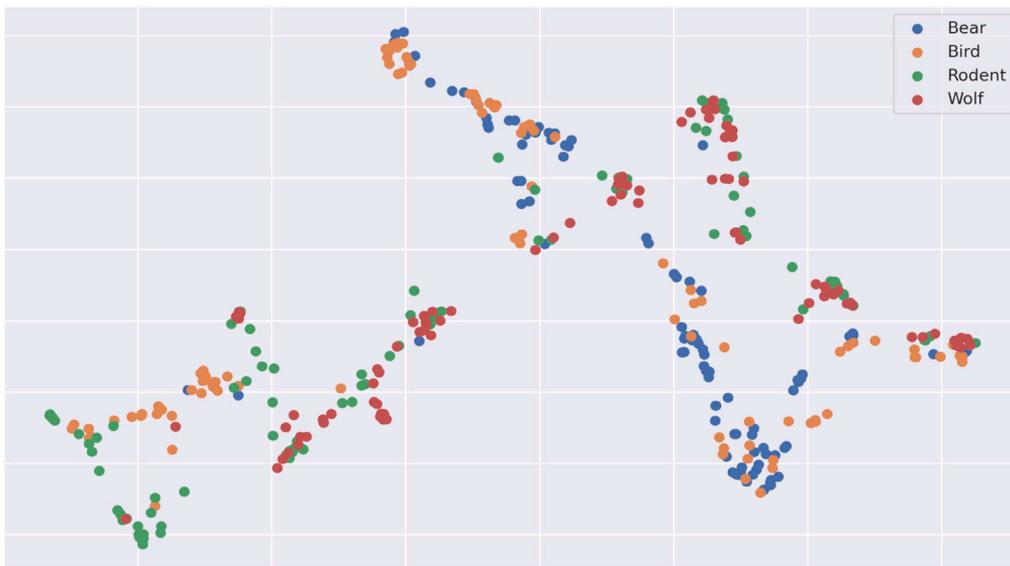

**Fig. 5.** Subsequent iterations of UMAP no longer cleanly separates the classes.

exhibit unique activity patterns or habitat preferences in response to certain environmental conditions (e.g., temperature, time of day). Following the isolation of the Mustelidae class, we proceeded to reapply the algorithm to the nine remaining classes, as illustrated in Fig. 3b. This process involves progressively eliminating the most distinct class and then rerunning the UMAP projection on the remaining data. As a result, we successfully separated the data into distinct categories, namely, "Fox", "Feline", "Farm Animal", and "Boar", as depicted in Fig. 4. However, the remaining classes exhibited a lack of clear separation based on metadata, as evident in Fig. 5.

Finally, we can gain valuable insights by examining samples from various classes, shedding light on the obstacles the neural network must surmount to classify them accurately. Fig. 6a illustrates a deer partially concealed by its intricate coat, while Fig. 6b highlights the complications arising from flash usage in low-light conditions. Additionally, we confront the challenge of handling partial images, as evident in Fig. 6c and Fig. 6d. While these images indeed pose challenges, they are not the most demanding in our dataset. Some images present only faint traces of animal features, shown in Fig. 7a, or are so severely blurred that discerning any details becomes a formidable task, as seen in Fig. 7b.

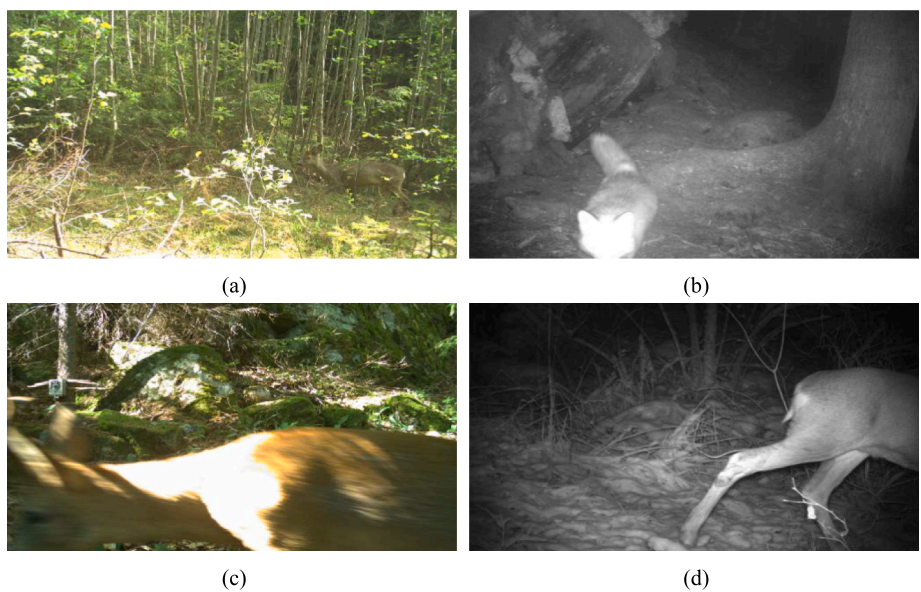

**Fig. 6.** Image variety and challenges

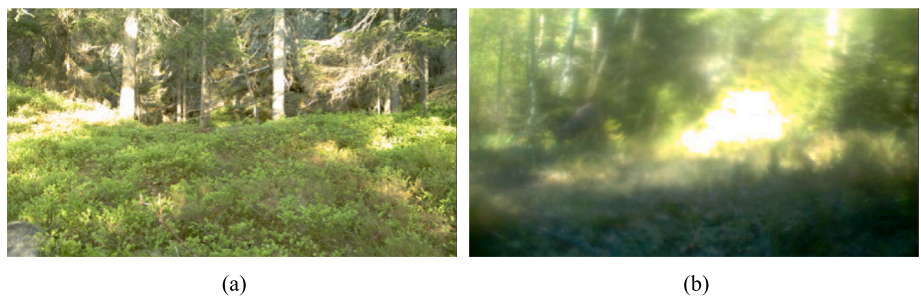

**Fig. 7.** Challenging images

### 3.2. Metadata

Accessing metadata presented its own set of challenges. While some metadata was readily available in the base downloadable file from the website or indirectly accessible through foreign key references, we encountered missing temperature data in the majority of the samples. In this section, we detail our approach to addressing this issue, including tackling the circular nature of date-time and explaining our rationale for incorporating location information. Additionally, we delve into the handling of implicit metadata.

#### 3.2.1. Temperature
Temperature is a critical variable in species distribution modeling. Integrating temperature information with camera trap data allows researchers to create more accurate models predicting where specific species are likely to be found based on environmental conditions. To estimate missing temperatures, we employed the Norwegian Meteorological Institute's Frost API[7]. Using the image's capture time, latitude, and longitude, we retrieved data from the nearest weather station within ±24 hours of the image. It is worth noting that some images lacked corresponding weather station readings within this time frame, and temperature fluctuations during the day were not considered. For any remaining missing values post-Frost API usage, we implemented a 2D vector approach. The first value indicated the validity of the temperature reading: 1 for valid and 0 for invalid. This method helped the network discern when to trust or ignore the temperature value.

#### 3.2.2. Datetime
Date and time values were encoded as a compact 67-dimensional one-hot vector. This approach was selected to accommodate the circular nature of time representation in neural networks. By allocating 12 dimensions for the month, 31 for the day, and 24 for the hour, we maintained the original date information without introducing discontinuities at year ends. While sine curves were initially considered due to their circular nature, concerns arose regarding equivalent values occurring during seasonal transitions (e.g., spring and fall) and specific times of day (e.g., dawn and dusk).

#### 3.2.3. Location
We included latitude and longitude with the suspicion that animals might occupy distinct regions, especially when considering time. This data, accessible through the web scraper, has limited precision — it is accurate only within approximately a kilometer of the pin's location. However, we acknowledge that this level of precision may not fully capture the extensive movements of animals, which often roam across areas larger than 1 kilometer. It is important to note that our location information is approximate, and the data's granularity may not fully represent the animals' broader range of activity. The location information was just the latitude/longitude pair for each camera trap. The positional data were included in the metadata vector and concatenated with other features. To standardize the values, min-max normalization was applied, considering that most geographical coordinates in Norway fall within the range of approximately 58 to 71 degrees latitude and 4 to 30 degrees longitude.

---
[7] https://frost.met.no/index.html

*3.2.4. Implicit metadata*

We supplemented the metadata with what we termed *scene attributes*, which were obtained using existing deep learning models detailed in (Zhou et al., 2017), which describe the visual content of images, identifying elements such as "leaves", "clouds", "trees", "mountain", and so forth. These models extracted 102 scene attributes from each image, which were subsequently stored alongside other image metadata. In addition to scene attributes, the pre-trained models from the Places365 dataset (Zhou et al., 2017) offered scene recognition capabilities, which were the dataset's primary focus. While scene attributes proved more beneficial in enhancing classification results, scene recognition still contributed to accuracy improvements in our ablation studies. Consequently, many of our tested models incorporated both scene attributes and scene descriptors to provide additional context for species classification. We consider implicit scene attributes as metadata derived from images that offer valuable insights into environmental conditions and habitat characteristics, enriching our understanding of wildlife observations. Incorporating scene attributes enhances the depth and comprehensiveness of our analysis, facilitating a holistic interpretation of wildlife observations and habitat dynamics as evident by the performance gains achieved in this study. Certain animals exhibit preferences for specific geographical features. For example, goats may prefer mountainous regions over deer. By incorporating a parameter indicating a high likelihood of the current image representing a "mountain" scene, the network is encouraged to predict goats, especially in ambiguous and hard-to-tell cases. Similarly, attributes are associated more strongly with certain species, reflecting specific visual characteristics. Our aim is for the CNN to focus on the visual traits of the target species from image data while learning about the habitat or scenery surrounding the animal from the scene attributes.

*3.3. Ablation study*

The intent of ablation study was to see how well metadata alone can differentiate between animal classes; therefore, we used a fairly simple, fully connected model, with the input layer matching the number of metadata features in the current ablation test. This layer's size varied based on the available features. The two hidden layers consisted of 128 and 64 nodes, respectively, followed by a final classification layer with output nodes corresponding to the number of classes in the current run. In this context, the models are solely provided with metadata (with no images). Ideally, we would aim to investigate which specific metadata factors have the most significant impact on prediction accuracy and which groupings of animals benefit the most from particular metadata attributes. These additional experiments enable us to determine whether species can be classified using metadata features alone. If successful, this supports the notion that metadata is valuable and will likely enhance the network's performance when combined with images. However, this exploration rapidly becomes impractical due to the sheer complexity of the task. Metadata can be categorized into several components, including datetime, temperature, position, scene attributes, and scene descriptors. Simultaneously, we are dealing with 13 classes that warrant investigation. When considering all possible combinations of pairs, triples, quadruples, and so on of animals, coupled with permutations of metadata, the number of potential combinations reaches a staggering 253,518 – a computationally unmanageable figure. As a pragmatic solution, we opted to streamline the study by reducing the number of animal classes to 9 (Fox, Deer, Mustelidae, Bird, Lynx, Cat, Sheep, Rodent, Wolf). Additionally, we merged the relatively concise position and temperature information into a single combined vector labeled as "*pos_temp*". This strategic simplification effectively reduces the number of combinations to a manageable 7,529.

*3.3.1. Counting method*

To identify the most effective metadata feature(s) for classifying animal species. We developed a counting method that evaluates the performance of different metadata features (or combinations thereof) across multiple classification trials. The analysis is divided into three parts: identifying the best single metadata feature, the best pair of metadata features, and finally, the best trio of metadata features.

For the first analysis, each metadata feature was used to classify different combinations of species (e.g., combinations of 2 species, 3 species, 4 species, and so on, up to 9 species). Each possible combination of animal species is referred to as a "classification trial". The metadata feature that achieved the highest accuracy in any given classification trial was awarded a point. This process was repeated across numerous trials, each time with different species combinations, to ensure robustness. After conducting all possible classification trials (total class combinations: 7,529), we tallied the points to determine how often each metadata feature was the top performer. This same process was conducted for pairs and trios of metadata features to distinguish between combinations of species. The counting method is a straightforward yet powerful tool for determining which features are consistently the best predictors, though it does not account for the magnitude of performance differences between features.

*3.3.2. Borderline synthetic minority oversampling technique (SMOTE)*

To mitigate the challenges posed by imbalanced datasets, we employed the Borderline Synthetic Minority Oversampling Technique (Borderline SMOTE) (Han et al., 2005), enabling us to augment the training data used for the network while keeping the validation and testing data unaltered. The plain SMOTE algorithm has a weakness in that it may generate synthetic samples anywhere in the higher dimensional space. Borderline SMOTE aims to improve this. By only generating synthetic samples on the boundary region between classes, the network gets more hard to tell samples, which should provide more benefit during training. It takes the set of all samples in the minority class class $P = \{p_1, p_2, \ldots, p_{pnum}\}$ and majority class $N = \{n_1, n_2, \ldots, n_{nnum}\}$. The next step is to count the number of majority class samples among the $m$ nearest neighbors of a given point $p_i$. This number $m'$ could be any number between $0 \leq m' \leq m$. If $m' = m$ the point is marked as noise and ignored going forward. If $m/2 \leq m' \leq m$ the minority sample is marked as a "DANGER" point, as it is likely to be misclassified, as 50% or more of its nearby neighbors are of the majority class. After all minority samples are checked, new synthetic samples are generated based on the minority samples in the DANGER group.

*3.4. Our metadata augmented models*

Our models are built upon the ResNet50 architecture, with specific modifications tailored to integrate metadata seamlessly. Although alternative models demonstrated slightly superior performance compared to ResNet50, we opted for ResNet50 for several reasons. Firstly, the design of the ResNet50 model provides a more straightforward mechanism for fusing metadata with the convolutional blocks, facilitating an intuitive incorporation of metadata into the model. Secondly, ResNet50's residual blocks are designed to learn hierarchical feature representations effectively. This hierarchical feature extraction is crucial for combining diverse data sources like images and metadata, as it ensures that both low-level and high-level features are well-captured and utilized.

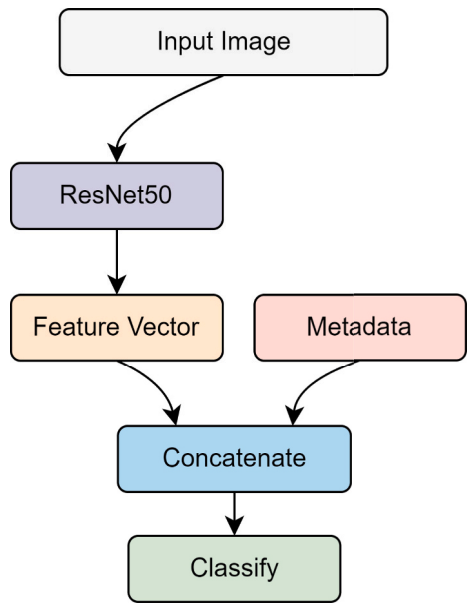

**Fig. 8.** Our late fusion approach

*3.4.1. Late fusion model*

The simplest model conceptually employed in our study is the late fusion model. This model starts with a feature vector $\vec{v}_1 = ResNet50(x)$ obtained from ResNet50, representing the final features before the classification step. Additionally, we have another feature vector $\vec{v}_2 = \mathscr{M}$, which encapsulates metadata information. These two vectors are concatenated to form a new combined vector $\vec{v}_1 \oplus \vec{v}_2 = \vec{V}$. After concatenation of the vectors, we run the full vector $\vec{V}$ through three layers of a fully connected network:

$$\hat{y} = g_3\left(g_2\left(g_1\left(\vec{V}\right)\right)\right)$$

In this context, each of the functions $g_1$, $g_2$, and $g_3$ correspond to a fully connected linear layer followed by a ReLU activation function. The model can be visualized in Fig. 8.

*3.4.2. Early fusion model*

Another broad category of fusion is early fusion, also known as feature fusion networks. These networks aim to integrate various data modalities early in the process to enhance feature extraction. When focusing on early fusion, the question of how to fuse metadata becomes a central consideration. In our model, we implement metadata fusion at three key junctures within the bottleneck block of the Residual Net. This fusion is achieved by element-wise multiplication of the metadata vector with the feature map generated by each convolution block. To ensure compatibility, the metadata is passed through a linear layer with the same output shape as the feature map of the convolution layer. Subsequently, a Sigmoid function is applied to the metadata to introduce additional non-linearity into our network. Mathematically, each block of the Residual Net has been modified like this:

$$out = in + g_3(g_2(g_1(c(in) \times f_1(\mathscr{M})) \times f_2(\mathscr{M})) \times f_3(\mathscr{M}))$$

where $f_1$, $f_2$, and $f_3$ represent the nonlinear functions that enable the metadata to fit the shape of the feature maps generated by the convolutions. The sign $\times$ here implies a layer-wise multiplication between the feature map and the metadata. The different $g$'s represent normalization steps and nonlinear layers through which the convolution is passed and are present in a normal ResNet50 architecture as well. Finally, $c$ represents the first convolution done in each block. Fig. 9 provides an overview of our early fusion approach.

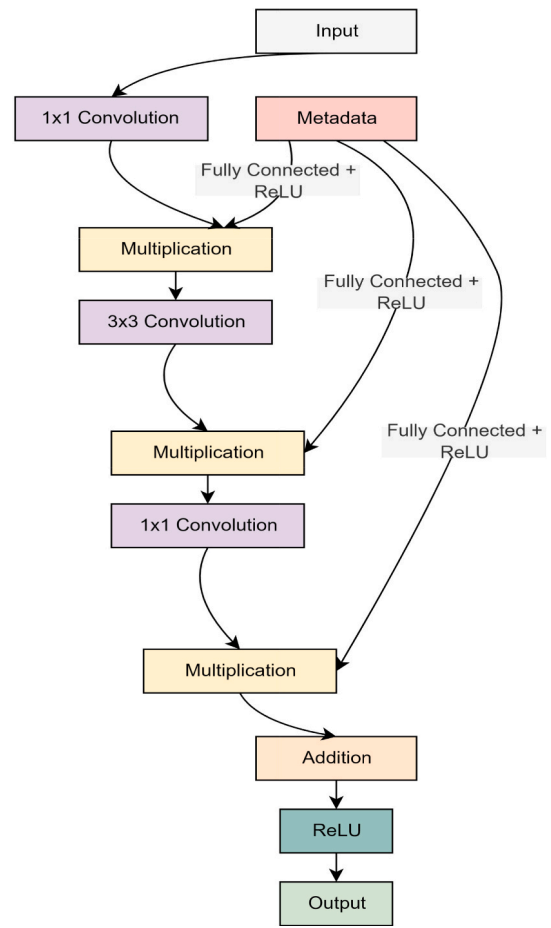

**Fig. 9.** Our early fusion approach

*3.4.3. Modified channel block attention module (MCBAM)*

The utilization of a channel block attention mechanism (CBAM) is motivated by its potential to enhance feature extraction and facilitate selective attention to informative regions of images, and we test whether it can also adapt to contextual information (metadata). We wanted to employ an attention-based model alongside traditional deep neural networks for comparative analysis, to see how well the attention mechanism measures up when compared against traditional approaches. CBAM was also chosen due to its compatibility with modifications. The idea was that this approach might enable the model to effectively discriminate between different animal species amidst complex backgrounds, leading to improved classification accuracy. Before we discuss our modifications, we need to discuss the mechanics of the attention module. At a high level, the attention mechanism employed by Woo et al. (2018) is demonstrated as follows: we multiply the incoming feature map by a channel attention vector and a block attention module. Mathematically written as:

$$F' = M_c(F) \otimes F,$$

$$F'' = M_s(F') \otimes F' \quad (1)$$

where $\otimes$ represents an element-wise multiplication. $M_c$ (channel attention map) and $M_s$ (block attention map) are defined as per the mathematical derivations from Woo et al. (2018):

$$M_c(F) = \sigma(MLP(AvgPool(F)) + MLP(MaxPool(f)) \quad (2)$$

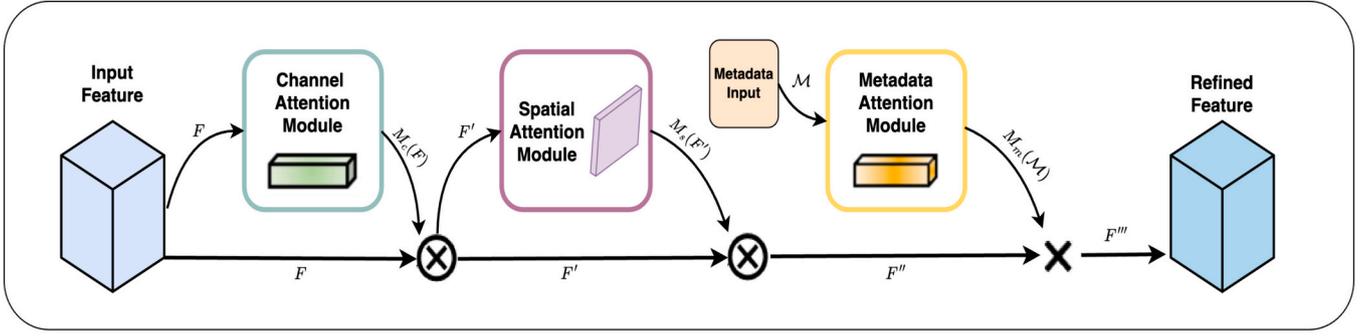

**Fig. 10.** Modified CBAM architecture

$$M_s(F) = \sigma\left(f^{7\times 7}([AvgPool(F); MaxPool(F)])\right) \quad (3)$$

where $\sigma$ represents the sigmoid function, and $f^{7\times 7}$ denotes a convolution of the input by a 7 by 7 kernel. $M_c(F)$ is derived by processing the global average-pooled and max-pooled features through separate multi-layer perceptrons (*MLP*). $M_s(F)$ is computed by applying a 7 by 7 convolutional operation to the concatenated average-pooled and max-pooled features. Both attention maps are then passed through a sigmoid activation function to obtain the final recalibration weights for channel-wise and spatial features.

We modified the Channel Block Attention Module by including an extra step after channel and block attention was applied to the network. We dub this last component "Metadata Attention". The overall structure of the network with our inclusion is depicted in Fig. 10.

More Formally, we can define the process as follows:

$$F' = M_c(F) \otimes F,$$

$$F'' = M_s(F') \otimes F',$$

$$F''' = M_m(\mathcal{M}) \times F'' \quad (4)$$

Do note that we multiply the metadata vector by the feature maps generated by $F''$. In line with methodologies used by Li et al. (2020) and Liu (2018). $M_m$ is then defined as:

$$M_m(\mathcal{M}) = \sigma(MLP(\mathcal{M})) \quad (5)$$

### 3.5. Experimental setup

#### 3.5.1. Baseline models

To thoroughly assess the impact of metadata on improving classification in WAC, we initiate our evaluation with baseline models. These serve the crucial purpose of providing performance for networks without metadata incorporation. Our baseline models encompass renowned architectures, including AlexNet (Krizhevsky et al., 2012), ResNet (He et al., 2016), InceptionV3 (Szegedy et al., 2014), EfficientNet (Tan and Quoc, 2020), and, notably, the Channel Block Attention Module (CBAM) (Woo et al., 2018).

#### 3.5.2. Evaluation metrics

In this study, we will use some commonly used metrics: precision, recall, $F_1$ score, false positive rate (*FPR*), and false negative rate (*FNR*), and overall accuracy. The formal definition for these are:

$$Precision = \frac{TP}{TP + FP}$$

$$Recall = \frac{TP}{TP + FN}$$

$$F_1 = \frac{2TP}{2TP + FP + FN}$$

$$FPR = \frac{FP}{FP + TN}$$

$$FNR = \frac{FN}{FN + TP}$$

These metrics can be conceptualized easily in a binary class:

$$Actual \begin{array}{c} Prediction \\ \begin{bmatrix} TN & FP \\ FN & TP \end{bmatrix} \end{array} \quad (6)$$

For a multi-class problem, the matrix becomes a bit more involved, but can be more generally summed up as:

$$Actual \begin{array}{c} Prediciton \\ \begin{bmatrix} TN & \ldots & TN & FP & TN & \ldots & TN \\ \vdots & \ddots & TN & \vdots & TN & \cdot\cdot\cdot & TN \\ TN & \ldots & TN & FP & TN & \ldots & TN \\ FN & \ldots & FN & TP & FN & \ldots & FN \\ TN & \ldots & TN & FP & TN & \ldots & TN \\ \vdots & \cdot\cdot\cdot & TN & \vdots & TN & \ddots & TN \\ TN & \ldots & TN & FP & TN & \ldots & TN \end{bmatrix} \end{array} \quad (7)$$

In essence, everything not in the row or column of a specific class is a true negative. Everything in the column of the class but not the row of the class is a false negative (wrong class predicted), and everything in the row of the class but not the column is a false positive (class predicted wrongly). This pattern holds for all classes in a multi-class problem.

To compute the overall accuracy of the model, we use:

**Table 1**
Metadata Predictors Scores: The table illustrates the best prediction results for m classes when using n metadata types.

| Classes | Features used | Acc | κ |
|---|---|---|---|
| 4, 6 | Scene attributes | 0.948 | 0.894 |
| 6, 12 | Position and temperature, Scene attributes | 0.982 | 0.945 |
| 4, 6 | Places, Position and temperature, Scene attributes | 0.967 | 0.932 |
| 6, 12 | Datetime, Places, Position and temperature, Scene attributes | 0.989 | 0.964 |
| 3, 4, 6 | Scene attributes | 0.87 | 0.779 |
| 3, 4, 6 | Position and temperature, Scene attributes | 0.869 | 0.782 |
| 3, 4, 6 | Datetime, Places, Scene attributes | 0.866 | 0.775 |
| 3, 4, 6 | Datetime, Places, Position and temperature, Scene attributes | 0.878 | 0.796 |
| 2, 3, 4, 6 | Scene attributes | 0.696 | 0.552 |
| 3, 4, 6, 12 | Position and temperature, Scene attributes | 0.731 | 0.603 |
| 3, 4, 6, 12 | Datetime, Position and temperature, Scene attributes | 0.729 | 0.614 |
| 3, 4, 6, 12 | Datetime, Places, Position and temperature, Scene attributes | 0.746 | 0.63 |

$$\text{Overall Accuracy} = \frac{1}{N}\sum_{i=0}^{N} k_i$$

where

$$k_i = \begin{cases} 1 & \text{if } \hat{y}_i = y_i, \\ 0 & \text{otherwise} \end{cases}$$

Finally, because of the imbalanced nature of our dataset, it can be useful to include a metric sensitive to prediction accuracy accounting for class imbalance. We will be using Cohens kappa score, proposed by Cohen (1960). Cohen kappa score measures the agreement between two predictors who classify $N$ items into $C$ mutually exclusive classes. To find this agreement, we need to first find the probability of our two predictors predicting identically by random chance $p_e$:

$$p_e = \frac{1}{N^2}\sum_{k=1}^{C} n_k^{(1)} n_k^{(2)}$$

Where $n_k^{(i)}$ is the number of times predictor $i$ predicted class $k$.

$P_o$ is the observed agreement between samples. Given some observed response matrix $M$:

$$M = \begin{bmatrix} x_{1,1} & \ldots & x_{1,C} \\ \vdots & \ddots & \vdots \\ x_{C,1} & \ldots & x_{C,C} \end{bmatrix}$$

$p_o$ is given as:

$$p_o = \frac{\sum_{i=1}^{C} x_{i,i}}{\sum_{i=0}^{C}\sum_{j=1}^{C} x_{i,j}}$$

Finally, we can use $p_e$ and $p_o$ to find the overall kappa score:

$$\kappa = \frac{p_o - p_e}{1 - p_e}$$

### 3.5.3. Oversampling and augmentation

Oversampling was performed using PyTorch "WeightedRandomSampler". The augmentation of the over-sampled images was performed with an augmentation pipeline using the Albumentations python package (Buslaev et al., 2018). Each augmentation technique had specific probabilities and limits:

- **Horizontal Flipping:** $P = 0.5$
- **Rotation:** $-45° \leq \theta \leq 45°$, $P = 1$
- **Color Jitter:** Brightness, contrast, hue, and saturation $\pm 0.1$, $P = 1$
- **Dropout:** Size $= 32 \times 32$, Holes $= 8$, $P = 1$

### 3.5.4. Training specifics

We divided the dataset into 90% training and 10% testing data. Within the training set, we performed an additional split, allocating 90% for training and 10% for validation. All models were pretrained on ImageNet and underwent training for 50 epochs and exhibited convergence. Image dimensions were standardized to 512 pixels. The models were developed and trained on a Linux system with an Intel i9 12900KF, 128 GB RAM, and an RTX3080-Ti. Initial weights were randomly set, and the optimizer's initial learning rate was $1e-3$, decreasing by an order of magnitude every seven epochs over 25 total epochs. Mini-batches of 64 samples were used, and each epoch included validation on 10% of the test samples. If the model's performance deteriorated, it was reverted to its best-performing iteration. Finally, the model was evaluated on the remaining 10% of the data initially set aside.

## 4. Results and discussion

In this section, we provide a thorough overview of the results from our metadata-augmented models for wild animal classification, alongside the performance of baseline models without incorporating any metadata. We begin by discussing the outcomes of our ablation study, where our models were exclusively provided with metadata for wild animal classification without utilizing any images.

### 4.1. Results of ablation study

The purpose of the ablation study was to find whether metadata could tell the difference between the animals at all. These experiments establish a baseline for the performance of metadata in isolation. This is crucial as it sets expectations for the incremental value added by metadata when combined with visual data. Knowing the baseline performance of metadata alone allows for a clearer quantification of the improvements achieved by integrating image data. In this context, we opted to utilize species IDs instead of the corresponding species names. The species IDs were as follows: 0 for "Fox", 1 for "Deer", 2 for "Mustelidae", 3 for "Bird", 4 for "Lynx", 5 for "Cat", 6 for "Sheep", 7 for "Squirrel", 8 for "Rabbit", 9 for "Rodent", 10 for "Cattle", 11 for "Boar", 12 for "Wolf", and 13 for "Bear". Our analysis focused on identifying the best prediction results for $m$ classes by using $n$ types of metadata. With fewer classes, the analysis is simpler and more straightforward, enabling researchers to easily interpret the results and understand the relationships between classes and metadata features. The results, as presented in Table 1, revealed that "Scene attributes" information emerged as the most influential single feature for enhancing prediction accuracy. Additionally, as we increased the number of included features, we observed a corresponding performance improvement. Moreover, the experiments demonstrate that metadata alone becomes insufficient as the number of classes increases, highlighting the need for fusion strategies that combine metadata with visual data.

However, the average performance of various features is not straightforward. We can better understand this relationship by examining the "winner" when comparing the performance of n predictors against each other across m classes. Fig. 11 illustrates the best predictor (s) for all combinations of all animals. "Scene attributes" emerged as the top single predictor, i.e we used only single metadata feature to try to classify different combinations of species/classes in Fig. 11a, and "SA" achieved the highest accuracy in many such experiments (we counted how many times it achieved the highest accuracy with different combinations of classes and this count is shown in the Y-axis) compared to "PI", "DT" or "P&T". And while "Scene attributes" emerge as the top single predictor, it is not among the top pairs, being surpassed by the combination of "Datetime" and "Places". Note that we are counting the "best predictor" for all combination of animals, and it does not consider the extent to which one predictor outperforms another. It is unclear whether "Scene Features" dominated as the sole feature or if other features closely followed its performance. Nevertheless, it is evident from Table 1 that accuracy generally improves with the inclusion of more features, indicating each metadata contributes value to animal feature prediction. Notably, these predictions are solely based on metadata, without image input.

As we introduce more classes, the predictive performance decreases, a trend depicted in Fig. 12. This decline is understandable because distinguishing between two classes is inherently easier than distinguishing among ten. However, even when considering the random guess factor, the kappa score also decreases. This implies that as the number of classes increases, the discriminative capability of metadata diminishes. To maximize the differentiation potential of metadata, we should concentrate on scenarios with as few classes as possible, ensuring that these classes are distinct when examined based on metadata.

### 4.2. Results of baseline and metadata augmented models

This section discusses the performance of the base models (trained solely on images) and our modified models (trained on both images and metadata). The complete results for all models are presented in Table 2. From the beginning of the table up to CBAM, the unmodified models are the baseline models. The models already demonstrate a reasonably high level of accuracy, making improvements relatively modest in percentage terms. Among the tested baseline models, InceptionV3 exhibits a slight performance advantage, with ResNet50 being a close second. It's worth noting that while only EfficientNetb3 is displayed, we tested all iterations of EfficientNet and retained the best-performing one for further evaluation.

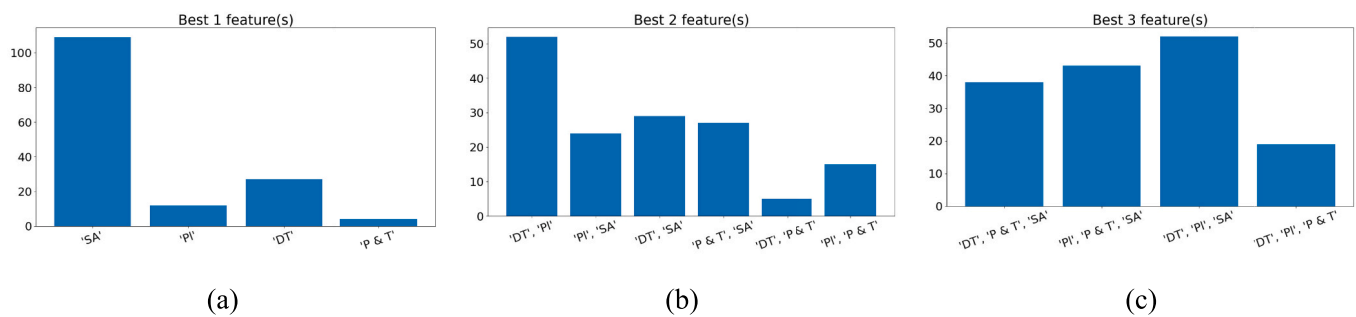

(a) (b) (c)

**Fig. 11.** The best *n* features to use to distinguish a set of *m* animals. (Y-axis shows the number of times metadata feature(s) was the best predictor (count) for all animal combinations, whereas the X-axis shows the metadata feature(s). "SA" for scene attributes, "Pl" for scene descriptors (Places), "DT" for datetime, and "P&T" for position and temperature data).

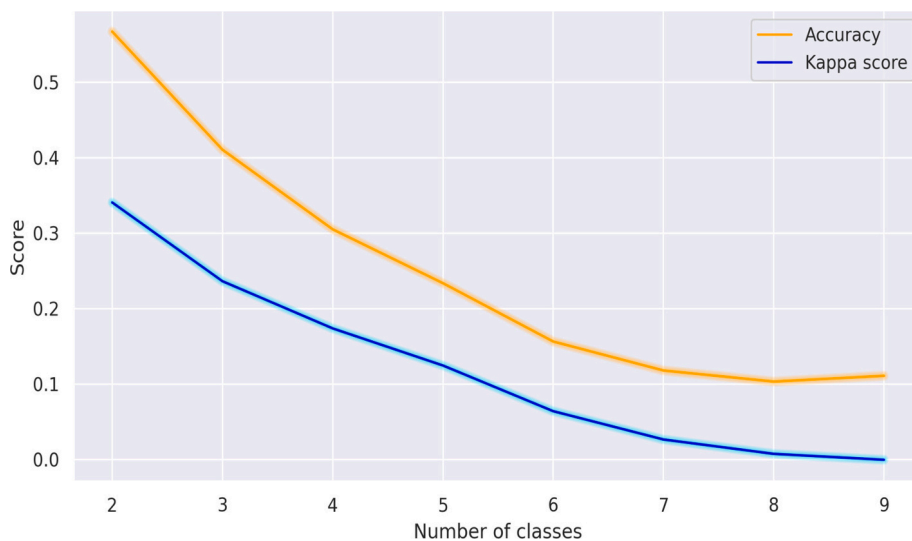

**Fig. 12.** Prediction score versus number of classes to distinguish

**Table 2**
Baseline Model Results: The table presents the performance evaluation of base models, trained solely on images, and modified models, trained on both images and metadata.

| Model | Accuracy | Precision | Recall | F1 | FPR | FNR | κ |
|---|---|---|---|---|---|---|---|
| ResNet18 | 0.966 | 0.953 | 0.964 | 0.958 | 0.003 | 0.036 | 0.959 |
| ResNet50 | 0.983 | 0.974 | 0.978 | 0.976 | 0.001 | 0.022 | 0.98 |
| AlexNet | 0.888 | 0.841 | 0.909 | 0.872 | 0.01 | 0.091 | 0.867 |
| EfficientNetb3 | 0.982 | 0.98 | 0.979 | 0.979 | 0.002 | 0.021 | 0.978 |
| InceptionV3 | 0.984 | 0.981 | 0.979 | 0.980 | 0.001 | 0.021 | 0.980 |
| CBAM | 0.805 | 0.688 | 0.789 | 0.720 | 0.017 | 0.211 | 0.768 |
| Late Fusion (Ours) | 0.987 | 0.986 | 0.98 | 0.983 | 0.001 | 0.02 | 0.984 |
| Early Fusion (Ours) | 0.989 | 0.984 | 0.986 | 0.984 | 0.001 | 0.014 | 0.987 |
| MCBAM (Ours) | 0.795 | 0.703 | 0.783 | 0.735 | 0.018 | 0.217 | 0.758 |

The results at the end of Table 2 indicate that two out of three models perform equally or better than our baseline models. The Modified Convolutional Block Attention Module (MCBAM) is the exception, as it does not benefit from metadata inclusion in terms of overall accuracy compared to other models tested. This could be due to several reasons. First, CBAM introduces additional layers and complexity, which increases computational overhead and resource requirements. This added complexity might not translate to better performance, particularly if the metadata is not effectively leveraged by the attention mechanisms. The increased computational demands can also make the model more difficult to train, potentially leading to suboptimal performance. Second, while CBAM and its modified version did outperform our Late and Early Fusion models with ResNet50 as the backbone or other traditional models, this might be due to CBAM's emphasis on specific channels and spatial locations. Such emphasis can lead to overfitting on the training set if the attention mechanism does not generalize well to the validation or test sets. This overfitting can result in poorer performance when the model encounters new data. Additionally, CBAM's integration of metadata may not be as seamless or effective as

Table 3

Comparison of the accuracy per class of the CBAM (Channel Block Attention Module) and MCBAM (Modified Channel Block Attention Module) models with counts for each species

| Class | CBAM | MCBAM | Count |
| --- | --- | --- | --- |
| Roe Deer | 0.829 | 0.792 | 56143 |
| Mustelid | 0.784 | 0.804 | 23332 |
| Fox | 0.746 | 0.699 | 22168 |
| Capreolinae | 0.818 | 0.877 | 14634 |
| Lepus | 0.855 | 0.798 | 12552 |
| Deer | 0.793 | 0.903 | 12262 |
| Bird | 0.823 | 0.777 | 10118 |
| Rodent | 0.679 | 0.719 | 6543 |
| Feline | 0.775 | 0.716 | 5640 |
| Farm Animal | 0.897 | 0.956 | 5477 |
| Wolf | 0.87 | 0.69 | 1256 |
| Boar | 0.846 | 0.786 | 393 |
| Bear | 0.545 | 0.667 | 115 |

Table 4

Comparison of the accuracy per class of the InceptionV3 and Early Fusion models with counts for each species

| Class | InceptionV3 | Early Fusion | Count |
| --- | --- | --- | --- |
| Roe Deer | 0.981 | 0.986 | 56143 |
| Mustelid | 0.982 | 0.99 | 23332 |
| Fox | 0.985 | 0.99 | 22168 |
| Capreolinae | 0.994 | 0.994 | 14634 |
| Lepus | 0.986 | 0.99 | 12552 |
| Deer | 0.988 | 0.997 | 12262 |
| Bird | 0.981 | 0.987 | 10118 |
| Rodent | 0.968 | 0.988 | 6543 |
| Feline | 0.991 | 0.989 | 5640 |
| Farm Animal | 0.993 | 0.995 | 5477 |
| Wolf | 0.992 | 0.992 | 1256 |
| Boar | 0.964 | 1.00 | 393 |
| Bear | 0.917 | 0.917 | 115 |

the straightforward fusion methods used with ResNet50. ResNet50's established robustness and the availability of pre-trained models likely provide a more stable and efficient foundation for incorporating metadata. In contrast, CBAM's attention mechanisms might struggle to balance the image features with the additional metadata, especially if the metadata is not optimally utilized. All of these reasons suggest that simpler and more established architectures like ResNet50, which are well-suited for metadata integration, can offer more reliable and superior performance in such tasks. However, MCBAM exhibits higher overall precision compared to CBAM but at the cost of reduced recall, implying a preference for lower false positives. We can evaluate this better by comparing CBAM and MCBAM's accuracy per class, shown in Table 3. We can observe that CBAM demonstrates superior performance in majority class classifications, while MCBAM exhibits slight improvements in minority class samples. The choice between these models depends on specific use-case requirements. Encouragingly, the inclusion of metadata does not inherently lead to a bias toward predicting the majority class.

The late fusion model slightly outperforms the best baseline models, but we are particularly interested in the early fusion model, which shows a slightly better performance. We observed that early fusion performs better than late fusion. We will now compare the performance per class of two models: the best-performing baseline model (InceptionV3) and the best-performing model from our metadata-augmented models (Early Fusion).

Table 4 indicates that the best baseline model performs better in only a few classes. This is encouraging, considering that we have tested only a limited number of architectures for metadata-augmented models. Other models will likely surpass the ones tested here, and it might even be worthwhile to design architectures from scratch, incorporating metadata-augmented feature extraction from the outset.

Relating our results to prior work in the field 2.2, Dhillon and Verma (2022) achieved 98.07% accuracy using DenseNet201 and ResNet101 on a camera trap dataset. Our best baseline model, InceptionV3, demonstrates comparable accuracy. Battu (2022) reported 73.09% accuracy with a noise-resilient approach on the Snapshot Serengeti and Panama-Netherlands datasets. Our baseline models show significantly higher accuracy, suggesting the effectiveness of our dataset and model selection. As far as metadata integration goes, Islam et al. (2023) achieved 87% accuracy with VGG16 and 86% with ResNet50 on multiclass classification of reptiles. Our metadata-augmented models surpass these figures, with the Early Fusion model achieving 98.9% accuracy. Finally, Sreedevi and Edison (2022) achieved 99.6% accuracy using a depth-wise separable convolution layer on the IWildCam dataset. While their approach focuses on convolutional efficiency, our method integrates additional contextual information through metadata, offering a different pathway to achieving high accuracy.

## 5. Conclusion and future work

This study showcased the importance of using metadata together with image data for effective classification of wild animals. Our results demonstrate that integrating metadata with image data significantly enhances the accuracy of wildlife classification models. The Early Fusion model, in particular, achieves superior performance compared to both our baseline models and those reported in previous studies. Our approach not only offers a robust solution for handling variations in image quality and lighting conditions but also sets a new benchmark for wildlife classification tasks by achieving high accuracy with metadata-only classification. To test the merit of using metadata, we conducted an ablation study that classified the animals by metadata alone; the results were a function of the number of classes in the dataset, decreasing with the increasing number of classes. However, they encouraged combining metadata with images. Notably, Scene Attributes, automatically extractable from any image, proved to be a powerful metadata feature. It is essential to emphasize that including metadata introduces more parameters, necessitating a rigorous evaluation of models with and without metadata. While other studies have demonstrated the efficacy of deep learning, our study is among the first in Wild Animal Classification to explore metadata utilization, and therefore, there is no established precedent to set our expectations. Thus, while data fusion has shown promise in other deep learning domains, its universal applicability in Wild Animal Classification requires further comprehensive research and experimentation.

We recommend further exploration of metadata-enhanced wild animal classification, particularly emphasizing the use of automated scene attribute and descriptor detectors. Moving forward, a promising avenue for further research involves the validation and adaptation of our methodology in diverse ecological contexts beyond Norway. Collaborative efforts with researchers and conservationists across different regions could facilitate the refinement and optimization of our approach to suit varying environmental conditions and wildlife populations. Moreover, exploring time encoding using sine and cosine components could be a promising avenue for addressing the issue of spring/fall indistinguishability. While this approach was considered later in the current project, time constraints prevented its implementation and testing. Future work may also explore alternative methods of metadata fusion could be investigated to enhance model efficiency and effectiveness.

**CRediT authorship contribution statement**

**Aslak Tøn:** Writing – original draft, Methodology, Formal analysis, Data curation. **Ammar Ahmed:** Writing – original draft. **Ali Shariq Imran:** Writing – review & editing, Supervision, Methodology, Conceptualization. **Mohib Ullah:** Writing – review & editing,

Supervision, Methodology. **R. Muhammad Atif Azad:** Writing – review & editing.

**Declaration of competing interest**

None.

**Data availability**

The implementation code used in this study is publicly accessible on GitHub[1] 1https://github.com/ammarlodhi255/metadata-augmented-neural-networks-for-wild-animal-classification. The repository contains scripts for data collection, processing, analysis, training, evaluation, and visualization, along with detailed documentation for usage, installation, and dataset preparation.

Model weights are available on Figshare[2] 2https://figshare.com/s/c4ca09789621053d5cb7. Additionally, important JSON files, which include species data and associated metadata, can be found here[3] 3https://figshare.com/articles/dataset/JSON_files/26832049?file=48793426.

The curated image dataset used in our main experiments is provided in two parts. The first part is the file named *images_part_1.zip* which can be accessed at this link[4] 4https://figshare.com/articles/dataset/images_part_1/26832043?file=48792940 and the second part is the file *images_part_2.zip* which can be found here[5] 5https://figshare.com/articles/dataset/images_part_2/26832475?file=48793543. To reconstruct the complete dataset, these files must be merged.